\documentclass[letterpaper]{article} % DO NOT CHANGE THIS
\usepackage{aaai2026}

\usepackage{times}  % DO NOT CHANGE THIS
\usepackage{helvet}  % DO NOT CHANGE THIS
\usepackage{courier}  % DO NOT CHANGE THIS
\usepackage[hyphens]{url}  % DO NOT CHANGE THIS
\usepackage{graphicx} % DO NOT CHANGE THIS
\usepackage{natbib}  % DO NOT CHANGE THIS AND DO NOT ADD ANY OPTIONS TO IT
\usepackage{caption} % DO NOT CHANGE THIS AND DO NOT ADD ANY OPTIONS TO IT
\usepackage{algorithm}
\usepackage{algorithmic}
\usepackage{amsmath}
\usepackage{xcolor}
\usepackage{amsfonts}

\usepackage{newfloat}
\usepackage{listings}
\DeclareCaptionStyle{ruled}{labelfont=normalfont,labelsep=colon,strut=off} % DO NOT CHANGE THIS
\floatstyle{ruled}
\newfloat{listing}{tb}{lst}{}
\floatname{listing}{Listing}
\usepackage{xcolor}
\newcommand{\answerYes}[1]{\textcolor{blue}{#1}} 
\newcommand{\answerNo}[1]{\textcolor{teal}{#1}} 
\newcommand{\answerNA}[1]{\textcolor{gray}{#1}}

\title{POLAR:\\ A Per-User Association Test in Embedding Space}
\author{
    Pedro Bento, Arthur Buzelin, Arthur Chagas, Yan Aquino, Victoria Estanislau, \\Samira Malaquias, Pedro Robles Dutenhefner, Gisele L. Pappa, Virgilio Almeida, Wagner Meira Jr.
}
\affiliations{
    Universidade Federal de Minas Gerais (UFMG), Belo Horizonte, Brazil

    \{pedro.bento, arthurbuzelin, arthurchagas, yanaquino, victoria.estanislau, samiramalaquias, virgilio, glpappa, meira\}@dcc.ufmg.br, pedroroblesduten@ufmg.br
}

\begin{document}

\maketitle

\begin{abstract}
Most intrinsic association probes operate at the word, sentence, or corpus level, obscuring author-level variation. We present \emph{POLAR} (Per-user On-axis Lexical Association Report), a per-user lexical association test that runs in the embedding space of a lightly adapted masked language model. Authors are represented by private deterministic tokens; POLAR projects these vectors onto curated lexical axes and reports standardized effects with permutation $p$-values and Benjamini--Hochberg control. On a balanced bot--human Twitter benchmark, POLAR cleanly separates LLM-driven bots from organic accounts; on an extremist forum, it quantifies strong alignment with slur lexicons and reveals rightward drift over time. The method is modular to new attribute sets and provides concise, per-author diagnostics for computational social science. All code is publicly available at \url{https://github.com/pedroaugtb/POLAR-A-Per-User-Association-Test-in-Embedding-Space}.
\end{abstract}

\section{Introduction}

Measuring social associations in language technologies is often approached through intrinsic probes that quantify how strongly words relate to semantic categories in embedding spaces. Among such probes, association tests inspired by psychological instruments have become widely used to study how distributional representations encode stereotypes and domain semantics across tasks and models \citep{sun-etal-2019-mitigating,gonen-goldberg-2019-lipstick-pig,kurita-etal-2019-measuring}. These tests provide a compact, model-agnostic signal, by projecting embeddings onto interpretable lexical axes and ask whether observed differences are larger than expected under exchangeable labelings of the attribute sets.

At the same time, \emph{user embeddings} have surged in prominence across machine learning subfields. In recommender systems, neural collaborative filtering represents each person with a learned vector that captures stable preferences \citep{he10.1145/3038912.3052569}. In networked settings, node-embedding methods such as DeepWalk, node2vec, and GraphSAGE map users to points in a geometry that preserves social neighborhoods and behavioral regularities \citep{Perozzi_2014,10.1145/2939672.2939754,NIPS2017_5dd9db5e}. In dialogue, persona-conditioned models leverage profile representations to steer generation style and content \citep{zhang-etal-2018-personalizing}. Across these lines, a shared intuition emerges: low-dimensional user vectors act as condensed, reusable summaries of a person’s linguistic habits, social position, and topical affinities.

\begin{figure*}
    \centering
    \includegraphics[width=1\linewidth]{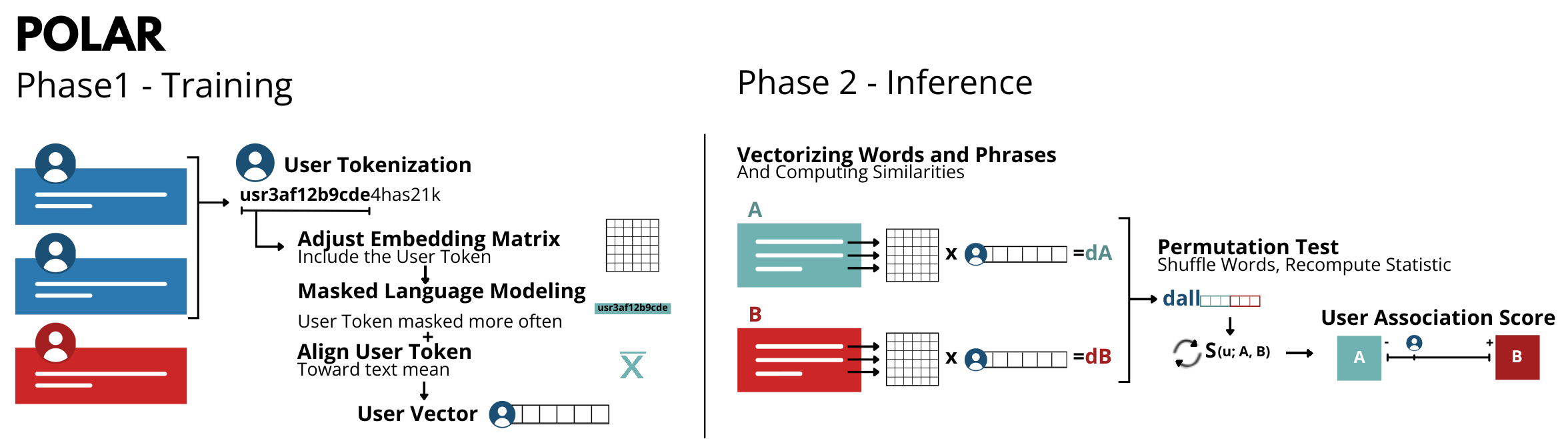}
    \caption{Overview of \emph{POLAR}, which first learns user vectors via masked language modeling with injected tokens (Phase~1), then computes association scores through cosine similarity and permutation testing against lexical attribute sets (Phase~2).}
    \label{fig:diagram}
\end{figure*}

Despite the ubiquity of intrinsic association tests and user-level embeddings, there is still no methodology that directly applies association testing to individual user vectors. Existing intrinsic probes overwhelmingly target words, sentences, or a single shared model space \citep{sun-etal-2019-mitigating,kurita-etal-2019-measuring}, so author-level associations are typically accessed only indirectly, via task-specific classifiers or by aggregating users into group summaries. This is limiting for computational social science: aggregation can obscure within-group heterogeneity, and in extreme cases even reverse or cancel effects that are consistently present at the individual level (the ecological fallacy) \citep{2bca0cca-e964-3217-ab80-6209b4628c61}. For instance, a corpus might exhibit a strong average association between “immigrants” and “crime” while actually mixing users who consistently use neutral or humanizing language with others who repeatedly deploy slurs; a single aggregate score cannot distinguish these trajectories.

Concretely, common “aggregate-probe” strategies collapse many authors into one representation, for example, computing a corpus-level WEAT/SEAT score in a shared embedding space, averaging sentence-level association scores across all posts, or pooling representations such as mean [CLS] over the dataset. These summaries answer a different question: they characterize the average association of the corpus, not the distribution of associations across authors \cite{wakefield2010spatial}. POLAR instead treats each author as the unit of analysis, producing per-user effect sizes $s(u;\mathcal{A},\mathcal{B})$ that retain heterogeneity and can reveal whether an association is widespread, concentrated in a minority, or shifting over a user’s posting history.

This gap motivates a method that treats each user vector as a first-class object for calibrated association testing. We propose \emph{POLAR}, a per-user lexical association test that operates in the same embedding space as the language model. Concretely, we hash each author to a private token, lightly adapt a masked language model so that the token’s vector summarizes that author’s linguistic context, and then compute standardized effect sizes by projecting the user vector onto curated semantic axes (e.g., \emph{X} vs.\ \emph{Y}) with Monte Carlo permutation $p$-values and false-discovery control. Because user and lexical embeddings cohabit the same space, associations are immediately interpretable in both sign and magnitude, revealing not only \textit{which side} of a semantic contrast a user falls on, but also \textit{by how much}, without requiring labels or external supervision.  POLAR thus enables: (i) user-token adaptation in pretrained language models, (ii) effect-size inference over lexical axes, and (iii) author-level diagnostic applications, bridging a key methodological gap in current probing paradigms. 

We demonstrate that POLAR surfaces author-level structure that aggregate probes miss. On a balanced bot–human Twitter corpus, POLAR recovers clear register differences across multiple axes without using labels at training time. On an extremist forum, POLAR quantifies strong positive alignment with slur lexicons and reveals rightward drift as users continue posting -- an interpretable, label-free indicator of progressive radicalization. Together, these findings show that (i) compact user vectors capture stable lexical preferences; (ii) association testing can be cleanly adapted to the author level; and (iii) the resulting scores provide actionable, statistically grounded diagnostics for sociotechnical analysis.

\section{Related Work}
Here we span three strands: (i) user representations and compact conditioning that learn concise per-user vectors; (ii) intrinsic association probes (e.g., WEAT/SEAT) applied to words, sentences, or shared spaces; and (iii) author-identified datasets on bots, toxicity, and political discourse. These threads motivate our contribution of applying calibrated association testing directly to individual user embeddings.

\subsection{User Representations in NLP and CSS}
User embeddings are now a standard primitive for personalization and social analysis, learned from text, interactions, or both \citep{pan2019social,kumar2019predicting,cecillon2021graph,pougue2023learning,cinus2025inference}. Dense user vectors have long supported personalization and social analysis, with early work learning multiview embeddings from text, networks, and profile signals to predict behavior and demographics \citep{benton-etal-2016-learning}. In dialogue, persona-based models encode speaker embeddings to capture style and audience effects across interactions \citep{li-etal-2016-persona}. Beyond prediction, user-level vectors illuminate audience effects and individual differences in community settings \citep{sepahpour2023using}. These works established that users can be represented as stable vectors, making it possible to analyze individual differences and audience effects in online communities.

\subsection{User Profiling and Bot Detection with Embeddings}
A substantial literature builds user representations from social media activity for profiling and detection tasks, including bot identification.
For example, \citet{alekseev2017word} study word-embedding-based user profiling in online social networks, and \citet{heidari2020deep} use deep contextualized embeddings for text-based user profiling to detect social bots on Twitter.
More broadly, user embeddings have been leveraged to separate bots from organic accounts using learned representations of language and behavior.
POLAR is complementary: instead of training a task-specific classifier as the primary object, it provides per-user association effect sizes with permutation $p$-values along interpretable lexical axes, which can then optionally be used as features for downstream models.

\subsection{Compact Conditioning for Personalization}
Recent work compresses user histories into low-dimensional vectors or soft prompts that efficiently steer language models without concatenating long input traces \citep{doddapaneni-etal-2024-user,10.1145/3711896.3737385}. Prompt tuning, prefix-tuning, adapters, and low-rank adaptation all demonstrate that a small, efficient per-user state can steer large language models \cite{li-liang-2021-prefix, lester-etal-2021-power, pmlr-v97-houlsby19a, hu2021loralowrankadaptationlarge}. These methods demonstrate the practicality of compact per-user state for conditioning and retrieval.

\subsection{Intrinsic Bias and Association Tests}
The Word Embedding Association Test (WEAT) introduced cosine-based association testing with permutation significance for static embeddings \citep{doi:10.1126/science.aal4230}. Subsequent variants extend to sentence encoders and highlight design sensitivities \citep{DBLP:journals/corr/abs-1903-10561,goldfarb2020intrinsic}. Prior work has noted that intrinsic bias scores can be unstable or only loosely connected to downstream harms \cite{blodgett-etal-2020-language}. These analyses, however, remain largely aggregate, targeting words, sentences, or model-wide shared spaces and treating the corpus or model as a single unit of analysis. As a result, they miss heterogeneity across individual authors and cannot distinguish whether a stereotype is uniformly encoded across users or concentrated in a subset of accounts. These concerns motivate complementary approaches that refine how such probes are applied, for instance by shifting attention from aggregate to individual-level embeddings.

In parallel, user-level embeddings are widely used for personalization and prediction in social media, including mental-health assessment \citep{amir2017quantifyingmentalhealthsocial}, lifestyle and motivation modeling \citep{Islam_Goldwasser_2021}, and unsupervised stance detection in polarized settings \citep{Rashed_Kutlu_Darwish_Elsayed_Bayrak_2021}, with broader surveys reviewing social-media-based user embedding methods \citep{ijcai2019p881}. These approaches typically learn low-dimensional user representations from posting histories or social graphs and then feed them into downstream classifiers or clustering pipelines, without quantifying calibrated associations with specific lexical axes. POLAR bridges this gap by treating individual user vectors as first-class targets of association testing, yielding author-level effect sizes and p-values along interpretable semantic frames rather than corpus-level surrogates.

\begin{figure*}
    \centering
    \includegraphics[width=1\linewidth]{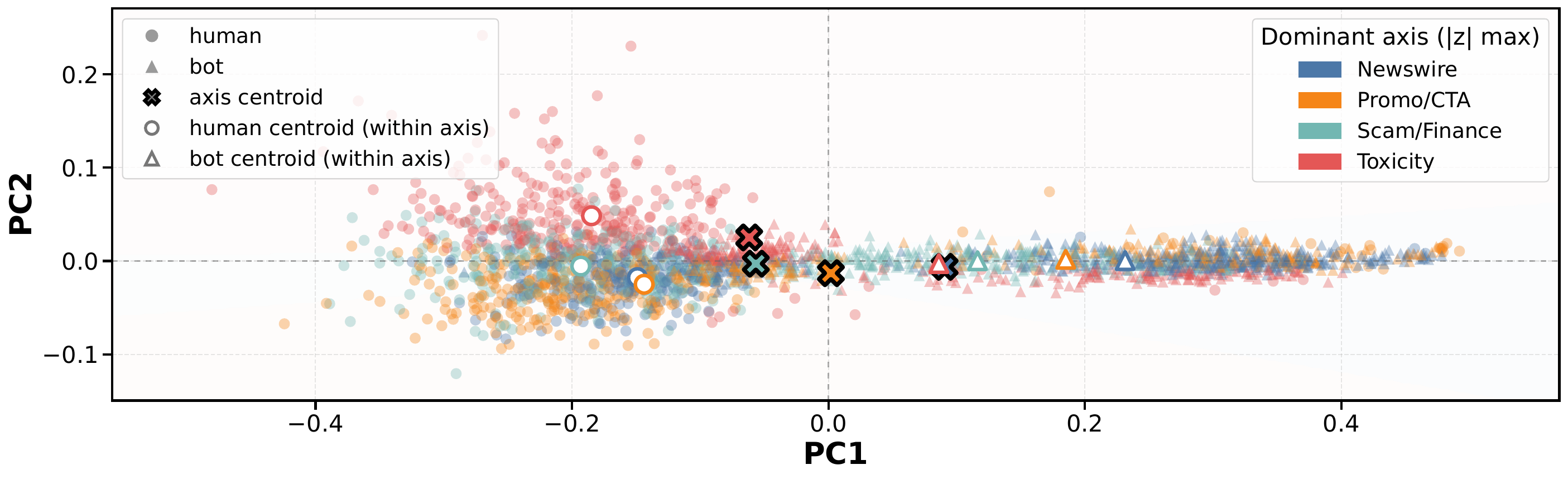}
    \caption{PCA projection of users in the 4D POLAR-axis score space. Marker shape encodes the gold label (bot vs.\ human), while color indicates the dominant axis (largest $|s|$ for that user). Centroids (black \texttt{$\times$}) summarize average positions; PCA is used for visualization, while classification is performed in the original axis space (Tables~\ref{tab:axis-bots}--\ref{tab:fox8-oof}).}
    \label{fig:bot-human-2d}
\end{figure*}

\subsection{Bots, Toxicity, and Political Discourse}
The fox8-23 benchmark documents LLM-powered bot accounts alongside humans \cite{Yang_Menczer_2024}, while hate-speech corpora from extremist forums such as Stormfront \cite{de-gibert-etal-2018-hate} provide author-identified data on hostile language. Beyond NLP, social science studies have documented how exposure and community norms shape polarization and radicalization \cite{doi:10.1073/pnas.1804840115, matias}. Hate/toxicity corpora with author identifiers provide a complementary lens on hostile language and community norms, while political-discourse datasets capture stance and polarization in the wild. Together they stress-test whether POLAR captures domain-general signals or domain-specific structure.

Despite advances in user embeddings, compact personalization, and intrinsic bias probes, no prior work has systematically adapted association testing to the level of individual users. Existing approaches remain focused on words, sentences, or aggregated representations, overlooking how associations may vary across individual authors. To our knowledge, this leaves an actionable gap: a method that treats user embeddings as first-class targets of calibrated association testing, yielding interpretable, statistically grounded, per-user scores rather than aggregate surrogates.

\section{Methodology}
\label{sec:method}

We introduce \emph{POLAR}: a per-user lexical association test run \emph{directly} in the embedding space of a masked-language model (MLM). The pipeline has two phases. First, we obtain a compact vector for each user by inserting a private, deterministic user token into their posts and lightly adapting the MLM so that the token vector summarizes that user’s typical linguistic context. Second, holding the model fixed, we compare each user vector to curated attribute word/phrase sets (e.g., \emph{gun safety} vs.\ \emph{gun rights}) and compute standardized, permutation-tested effect sizes with false-discovery control, as illustrated in Figure \ref{fig:diagram}.

\subsection{Attribute-Set Construction and Documentation}
\label{ssec:attr-construction}
Attribute pairs $(\mathcal{A},\mathcal{B})$ define the semantic axes probed by POLAR, so their construction is part of the method rather than an afterthought.
We build these sets by combining resources from prior association-test work (e.g., WEAT/SEAT-style lists) with domain knowledge and platform-specific lexicons (such as bot-disclaimer phrases or slur/jargon lists used in hate-speech research), and we apply simple sanity checks to keep the resulting axes interpretable and stable.
In practice, we balance list sizes when possible, favor high-frequency lexical items and short phrases to reduce tokenization drop-out, remove near-duplicates and overly ambiguous items, and record per-(user, pair) coverage statistics indicating how many attributes survive tokenization.

\subsection{Data Construction and User Tokens}
Let $\mathcal{D}=\{(u_i,x_i)\}_{i=1}^N$ denote posts $x_i$ authored by users $u_i$. Each user $u$ is mapped to a deterministic token
\[
t_u=\texttt{usr}\Vert \text{SHA1}(u)_{[:10]},
\]
added to the tokenizer vocabulary and \emph{prepended} to every post: $\tilde{x}_i=(t_{u_i}\Vert x_i)$. We lowercase, apply wordpiece tokenization (BERT-base-uncased by default), discard users with fewer than two posts, and optionally cap per-epoch posts per user to curb dominance by prolific accounts.

\paragraph{Training sketch}
Starting from \texttt{bert-base-uncased}, we resize the input embedding matrix $E\in\mathbb{R}^{V'\times d}$ to include the user tokens. Batches are balanced across users. We optimize the standard masked-LM loss with user-aware masking (higher mask probability on the user token) and a small alignment term that nudges the user vector $E[t_u]$ toward the average representation of the post’s non-user tokens. A short freeze–then–unfreeze schedule stabilizes user token learning; all hyperparameters are documented in the Appendix.

\subsection{POLAR in the Learned Space}
After training, each user $u$ is represented by the row-normalized embedding
\[
\hat{\mathbf{e}}_u=\frac{E[t_u]}{\|E[t_u]\|_2}\in\mathbb{R}^d.
\]
Attributes are specified as two finite sets $(\mathcal{A},\mathcal{B})$ of lexical items, each item being either a unigram or a phrase. We map any item $w$ to an embedding by averaging its wordpiece vectors under the \emph{trained} tokenizer and then $\ell_2$-normalizing; denote this map by $\phi:\text{vocab}\to\mathbb{R}^d$. Stack the resulting unit vectors into matrices
\[
A \;=\; \begin{bmatrix} \phi(a_1)^\top \\ \cdots \\ \phi(a_m)^\top \end{bmatrix}\in\mathbb{R}^{m\times d},
\qquad
B \;=\; \begin{bmatrix} \phi(b_1)^\top \\ \cdots \\ \phi(b_n)^\top \end{bmatrix}\in\mathbb{R}^{n\times d}.
\]

\paragraph{Statistic.}
POLAR follows the geometry of the original WEAT statistic, adapted to the user-embedding setting. For a given user $u$, we measure cosine similarities
\begin{align*}
d_A &= A\,\hat{\mathbf{e}}_u \in \mathbb{R}^{m}, \\
d_B &= B\,\hat{\mathbf{e}}_u \in \mathbb{R}^{n}, \\
d_{\text{all}} &= 
  \begin{bmatrix} d_A \\ d_B \end{bmatrix} \in \mathbb{R}^{m+n}.
\end{align*}
and define the standardized effect size
\begin{equation}
\label{eq:userweat}
s(u;\mathcal{A},\mathcal{B})
\;=\;
\frac{\operatorname{mean}(d_A)-\operatorname{mean}(d_B)}
{\operatorname{sd}(d_{\text{all}})}\,.
\end{equation}
The numerator can be written $(\bar{\mathbf{a}}-\bar{\mathbf{b}})\cdot \hat{\mathbf{e}}_u$, where $\bar{\mathbf{a}}=\tfrac{1}{m}\sum_{i=1}^m\phi(a_i)$ and $\bar{\mathbf{b}}=\tfrac{1}{n}\sum_{j=1}^n\phi(b_j)$; thus $s$ is a projection of the user vector onto the semantic axis $\bar{\mathbf{a}}-\bar{\mathbf{b}}$ rescaled by the empirical dispersion of all similarities for that user–pair. Standardization yields a unitless, scale-stable quantity that is comparable across pairs with different $(m,n)$ and internal variability.

\paragraph{Null, $p$-value, and exchangeability.}
The null hypothesis states that user $u$ exhibits no differential association with $\mathcal{A}$ versus $\mathcal{B}$. Under this null, labels on the $m{+}n$ entries of $d_{\text{all}}$ are exchangeable. We estimate a two-sided $p$-value by Monte Carlo permutation with $M$ random re-labelings (default $M{=}2000$). Writing $s_{\text{obs}}$ for the observed statistic and $s^{(k)}$ for the statistic at permutation $k$, the estimate
\[
\hat{p}
\;=\;
\frac{1+\sum_{k=1}^{M}\mathbb{I}\big(\lvert s^{(k)}\rvert \ge \lvert s_{\text{obs}}\rvert\big)}{1+M}
\]
uses add-one smoothing to avoid zero $p$-values under finite $M$ and to bound the Monte Carlo error. This test requires no distributional assumptions beyond exchangeability and remains valid for small lists as long as $m,n>0$.

\begin{figure*}
    \centering
    \includegraphics[width=1\linewidth]{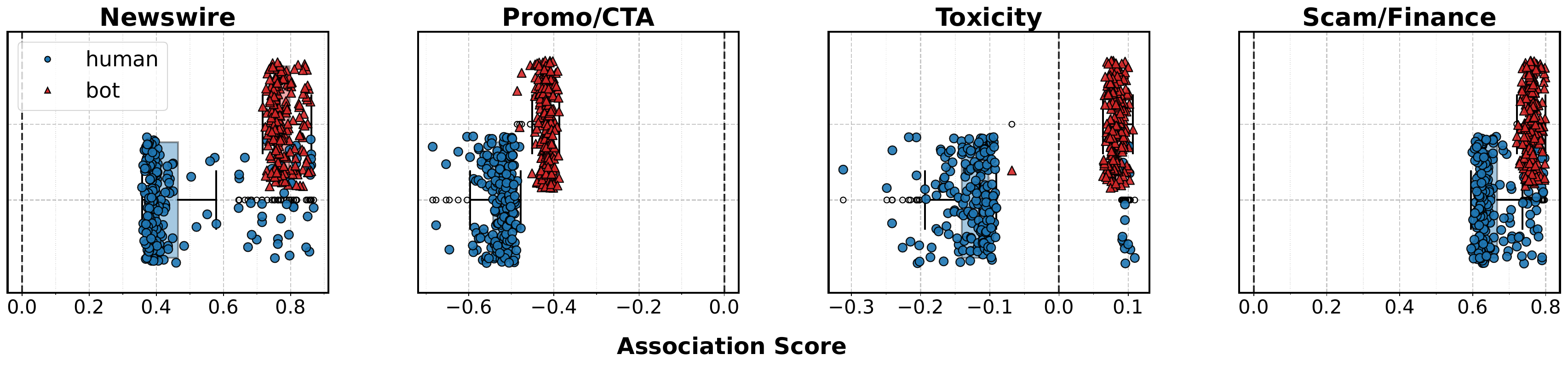}
    \caption{Box-and-swarm plots of per-user \emph{POLAR} scores $s(u;\mathcal{A},\mathcal{B})$ (y-axis; Eq.~\ref{eq:userweat}) for humans and bots across four lexical axes. Points are individual users; boxplots summarize median and interquartile range.}
    \label{fig:bot-human-dist}
\end{figure*}

\paragraph{Multiplicity Control.}
Because each attribute pair is tested across many users, we control the expected false discovery proportion \emph{per pair} via the Benjamini–Hochberg procedure at level $\alpha{=}0.05$, applied to the vector of permutation $p$-values $\{\hat{p}(u;\mathcal{A},\mathcal{B})\}_{u}$. Reporting both the raw statistic $s$ and the FDR-adjusted decision enables effect-size interpretation alongside inferential guarantees.

\paragraph{Interpretability and Benefits.}
Two properties make POLAR particularly useful for computational social science. First, user vectors and lexical items occupy the same embedding geometry; the sign of $s$ encodes direction along a named semantic axis, while its magnitude captures graded alignment. Analysts may therefore read results as \emph{which side} of a frame a user aligns with, and by how much, without training auxiliary classifiers. Second, inference is modular: adding or revising attribute sets does not alter the estimation or testing machinery, and phrase-level attributes are handled natively via subword averaging. In practice, we accompany per-user scores with basic diagnostics: realized coverage (how many attributes survived tokenization for a pair and user), a separability sanity check via $\cos(\bar{\mathbf{a}},\bar{\mathbf{b}})$, and explicit flagging of degenerate cases where $\operatorname{sd}(d_{\text{all}})=0$.

\subsection{Evaluation Protocol and Baselines}
\label{ssec:eval-protocol}
We evaluate discriminative performance on the fox8–23 bot–human benchmark using a one-layer logistic regression model trained on four distinct, label-agnostic feature sets. These include: (i) POLAR axes  --  per-user association scores $s(u;\mathcal{A},\mathcal{B})$ computed with labels held out; (ii) Mean CLS  --  the mean pooled [CLS] embedding of a user’s posts, reduced via PCA to 64 dimensions; (iii) Aggregated Sentence WEAT  --  post-level $s(x)$ scores using the same attribute sets as POLAR, averaged across posts per user; and (iv) Detox/Bot-speak proxies  --  average toxicity scores and simple behavior indicators such as URL, hashtag, mention, and call-to-action rates.

To avoid label leakage, we apply stratified 5-fold cross-validation. For (i), we select the top-$k$ axes within each training fold by ranking them via univariate AUROC. Logistic regression is then trained on those selected axes and evaluated on the held-out fold. We report out-of-fold (OOF) metrics, including AUROC with 95\% confidence intervals from 2,000 bootstrap resamples, as well as PR–AUC and Brier score. Paired bootstrap tests on OOF predictions assess whether increasing the number of axes (e.g., $k=2,3,4$) significantly improves performance over $k=1$. All POLAR computations  --  including axis statistics and permutations  --  remain fully unsupervised; only the logistic regression classifier is trained with labels.

Alongside discrimination, we report expected calibration error (ECE), computed on out-of-fold (OOF) probabilities with $10$ equal-width bins unless stated otherwise. ECE summarises the average gap between predicted confidence and empirical accuracy across bins; lower is better. We include ECE to show that the probabilities derived from single-axis POLAR scores can be used as risk-aware signals, not just for ranking.

\subsection{Reproducibility}
We fix stochastic seeds, store paths and hyperparameters in \texttt{meta.json}, and save the exact tokenizer (including user tokens). Unless noted otherwise, defaults are: sequence length $128$, batch size $128$, learning rate $5{\times}10^{-5}$, $4$ epochs, warm-up ratio $0.03$, gradient clip $1.0$, user-mask probability $0.30$, MLM mask $0.15$, alignment weight $0.2$, and $M{=}2000$ permutations. Mixed precision (bf16 when available, else fp16) is used automatically. Full training details are provided in the Appendix; the inference described above is unchanged by those choices.

\section{Datasets}

We evaluate POLAR on two corpora with user identifiers but distinct sociolinguistic regimes: (i) a balanced Twitter benchmark of LLM-powered bots and humans, used as a “vanilla” test of whether per-user lexical associations capture broad register differences; and (ii) a white-supremacist forum corpus with sentence-level hate labels, used to probe users’ stance and association with sensitive targets and policy frames.

\subsection{Bot–Human Twitter Benchmark (fox8–23)}
The fox8–23 collection comprises recent tweets from 1{,}140 accounts identified as AI-powered social bots and 1{,}140 human accounts, yielding a balanced bot–human sample at the \emph{user} level. Accounts were discovered and validated in a study of an LLM-driven botnet; the release provides tweet timelines for both groups.\footnote{We treat any dataset-provided account identifier as a user key for tokenization; raw screen names are never emitted by our pipeline.}

\paragraph{What we test with POLAR.}
This corpus serves as a sanity check that POLAR detects coarse stylistic divergences without supervision. We instantiate attribute pairs that contrast LLM-style “bot speak” against ordinary conversational language -- e.g., apology/safety disclaimers and instructional scaffolding (\emph{“I apologize”, “as an AI…”, “let’s break this down”}) versus colloquial, experiential, and interactional markers (\emph{“I think/feel”, “lol”, “gonna”, “in my experience”}). The hypothesis is that bots align more strongly with the former and humans with the latter; POLAR operationalizes this as a per-user standardized effect \(s(u)\) along each lexical axis with permutation \(p\)-values and BH–FDR control.

\subsection{Stormfront White-Supremacist Forum}
The Stormfront dataset contains 10{,}568 English sentences extracted from posts published between 2002 and 2017 on a white-supremacist forum. Sentences were annotated at sentence granularity (HATE vs.\ NOHATE, plus auxiliary labels) under detailed guidelines; crucially for our purposes, the release includes per-sentence \emph{user identifiers}, post positions, and subforum metadata, enabling reconstruction of user histories while shielding the original source via placeholder IDs.

\paragraph{What we test with POLAR.}
Here we use POLAR to quantify each user’s lexical alignment with sensitive topics and targets. Attribute pairs trace ideologically loaded frames and target-group semantics, contrasting derogatory or securitizing lexicons (\emph{e.g.}, outgroup slurs; “invasion”, “criminal aliens”) against neutral or humanizing counterparts (\emph{e.g.}, group self-references; “immigrants”, “refugees”, “equal rights”). Additional pairs cover policy frames frequently debated in such forums (immigration enforcement vs.\ reform, religious intolerance vs.\ freedom, gendered derogation vs.\ gender equality). POLAR yields per-user effect sizes and significance flags, surfacing heterogeneity within the community (users with strong positive/negative alignment on different axes) beyond aggregate hate/no-hate rates.

\paragraph{Preprocessing and inclusion.}
Across both corpora we lowercase, tokenize with the trained wordpiece model, and retain users with at least two posts (or sentences) to ensure stable vectors; each user is mapped to a deterministic hashed token and raw text is not required during testing.

\section{Results}
This section quantifies how POLAR surfaces meaningful linguistic variation at the author level. We first use the fox8--23 bot--human benchmark to verify that our method detects coarse stylistic divergences when the underlying language models differ sharply (LLM-generated vs.\ organic tweets). We then turn to Stormfront to probe ideological alignment and radicalisation. Unless noted, figures show standardized effect sizes $s(u;\mathcal{A},\mathcal{B})$ (Eq.~\ref{eq:userweat}) after Benjamini--Hochberg correction, with each dot representing a single user.

\subsection{Bot--Human Baseline}\label{ssec:bot-human-results}

Figure~\ref{fig:bot-human-2d} projects each user into the first two principal components of the four-dimensional POLAR score space.\footnote{Colours encode the user’s \emph{dominant} category, i.e., the category where $|s|$ is largest.} Users with similar association profiles concentrate near category centroids (black \texttt{$\times$}), suggesting that frames such as Promo/CTA or Toxicity capture stable register preferences. Clusters are elongated and overlapping: many accounts mix registers, and a non-trivial share of humans lie near bot-dominated regions along PC\textsubscript{1}, reflecting topic drift and short timelines.

To quantify separability, we train out-of-fold (OOF) logistic regressions on POLAR scores in two complementary ways: (i) one score at a time to assess the informativeness of individual axes (Table~\ref{tab:axis-bots}), and (ii) a compact top-$k$ model that selects axes within each training fold to test whether combining them helps (Table~\ref{tab:fox8-oof}). On the single-axis setting (users $N{=}2{,}277$), Scam/Finance vs.\ Daily Life and Newswire vs.\ Personal Experience each exceed AUROC~$0.95$, with strong PR--AUC and good probability quality (low Brier; moderate ECE), while Promo/CTA vs.\ Hedges remains informative and Toxicity vs.\ Civility is clearly weaker -- consistent with style and topical framing, not generic toxicity, driving the bot signal. The attribute anchor sets themselves are well separated in embedding space (cosine between positive/negative centroids $0.74$–$0.86$ across axes), supporting construct validity of the frames.

\begin{table}[t]
\centering
\small
\caption{Single-axis OOF performance on fox8--23 (one LR on each \emph{POLAR} score). Best two AUROC in \textbf{bold}.}
\label{tab:axis-bots}
\begin{tabular}{lcccc}
\hline
\textbf{Axis} & \textbf{AUROC} & \textbf{PR--AUC} & \textbf{Brier} & \textbf{ECE} \\
\hline
\textsc{Scam/Finance}      & \textbf{0.961} & 0.973 & 0.061 & 0.073 \\
\textsc{Newswire}          & \textbf{0.950} & 0.962 & 0.087 & 0.108 \\
\textsc{Promo/CTA}         & 0.915 & 0.930 & 0.116 & 0.068 \\
\textsc{Toxicity}          & 0.868 & 0.878 & 0.145 & 0.054 \\
\hline
\end{tabular}
\end{table}

A complementary view of raw score distributions appears in Figure~\ref{fig:bot-human-dist}: four box-and-beeswarm panels (one per axis) where humans (blue circles) and bots (red triangles) form two clouds with medians differing by more than one pooled standard deviation on every axis, and whiskers that rarely overlap -- making the separation visually apparent.

Finally, the top-$k$ OOF model (Table~\ref{tab:fox8-oof}) shows that adding axes beyond $k{=}1$ yields negligible differences: AUROC/PR--AUC remain essentially unchanged and calibration hardly improves. Paired bootstrap comparisons versus $k{=}1$ are not significant after Benjamini--Hochberg correction. Overall calibration is reasonable (ECE $\leq 0.108$), with Scam/Finance offering the best Brier ($0.061$).

\begin{table}[t]
\centering
\small
\caption{OOF performance on fox8--23 using a 1-layer logistic regression fed only with \emph{POLAR} axis scores. AUROC CIs from $2{,}000$ bootstrap resamples.}
\label{tab:fox8-oof}
\begin{tabular}{lccc}
\hline
\textbf{Features (axes)} & \textbf{AUROC [95\% CI]} & \textbf{PR--AUC} & \textbf{Brier} \\
\hline
POLAR ($k{=}1$) & 0.961 \,[0.952, 0.970] & 0.911 & 0.060 \\
POLAR ($k{=}2$) & 0.960 \,[0.951, 0.969] & 0.913 & 0.061 \\
POLAR ($k{=}3$) & 0.961 \,[0.952, 0.970] & 0.918 & 0.060 \\
POLAR ($k{=}4$) & 0.961 \,[0.952, 0.970] & 0.918 & 0.060 \\
\hline
\end{tabular}
\end{table}

Note the k{=}1 model reselects the top axis within each training fold by AUROC; consequently, its fold-varying choice can yield lower aggregate PR–AUC than the best fixed single-axis model in Table~\ref{tab:axis-bots}.

These results validate POLAR on a setting where linguistic differences are stark: without supervision, it exposes author-level effect sizes aligned with intuitive category semantics and yields diagnostic plots that make the separation visible.

\subsection{White-Supremacist Forum}
\label{ssec:stormfront-results}

\begin{figure}[!t]
    \centering
    \includegraphics[width=0.9\linewidth]{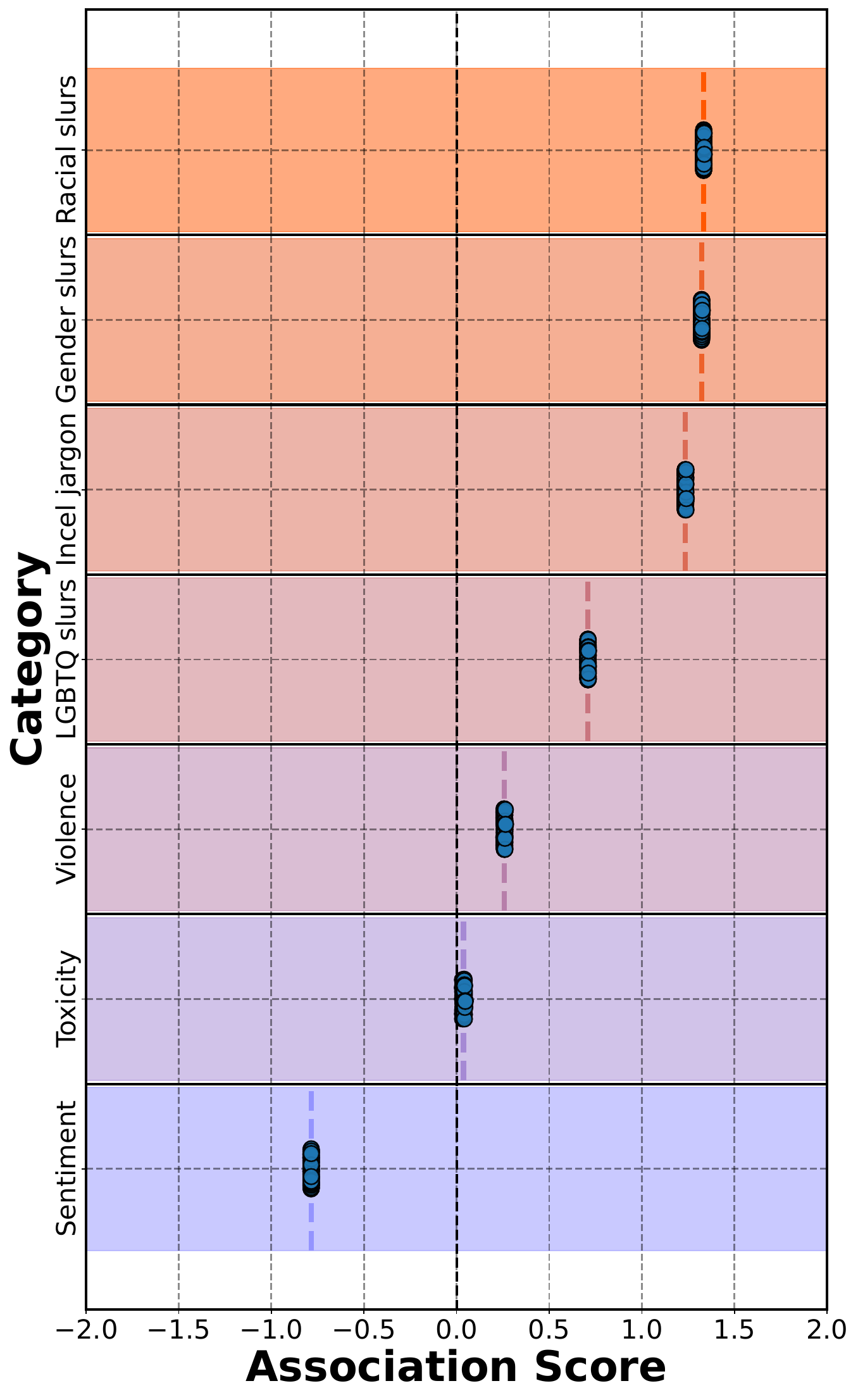}
    \caption{Stormfront per-user \emph{POLAR} associations on sensitive targets and policy frames.}
    \label{fig:forum-users}
\end{figure}

Figure~\ref{fig:forum-users} summarises POLAR scores for seven semantic axes in the Stormfront corpus ($N{=}2{,}082$). All slur-based categories -- Racial, Gender, LGBTQ, and Incel jargon -- cluster on the positive side, with means of $1.33$, $1.32$, $0.71$, and $1.24$, respectively; Violence shows a moderate positive mean ($0.26$), and Sentiment is negative ($-0.79$). The corresponding anchor centroids are also strongly separated (cosine $0.78$–$0.87$ across axes), indicating coherent attribute sets. Thus, in a community with an overt ideological leaning, POLAR pinpoints that bias numerically and visually, expliciting that the users embeddings sit closer to hateful or derogatory lexicons than to their neutral counterparts across every sensitive dimension we probe.

Because the forum lacks user-level gold labels, we report descriptive statistics for each axis: the mean standardized effect $s$ across users and the fraction significant after Benjamini--Hochberg at $\alpha{=}0.05$. Slur/jargon axes show large positive means with near-universal significance, while generic Toxicity and Violence exhibit small or non-significant shifts (Table~\ref{tab:hate-axis-descriptives}), indicating that \emph{domain-specific lexicons} -- not blanket hostility -- drive the embedding signal here.

\begin{table}[t]
\centering
\small
\caption{Stormfront: axis-level descriptives. \emph{Sig.\,BH} is the share of users significant after BH--FDR at $\alpha{=}0.05$ for that axis.}
\label{tab:hate-axis-descriptives}
\begin{tabular}{lcc}
\hline
\textbf{Axis} & \textbf{Mean $s$} & \textbf{Sig.\,BH} \\
\hline
\textsc{Racial Slurs vs.\ Neutral}       & 1.332  & 1.00 \\
\textsc{Gender Slurs vs.\ Respect}       & 1.320  & 1.00 \\
\textsc{Incel Jargon vs.\ Relationships} & 1.232  & 1.00 \\
\textsc{LGBTQ Slurs vs.\ Neutral}        & 0.707  & 1.00 \\
\textsc{Violence vs.\ Peace}             & 0.258  & 0.00 \\
\textsc{Toxicity vs.\ Civility}          & 0.033  & 0.00 \\
\textsc{Sentiment (neg.\ vs.\ pos.)}     & $-0.784$ & 1.00 \\
\hline
\end{tabular}
\end{table}

\begin{figure*}[t]
    \centering
    \includegraphics[width=1\linewidth]{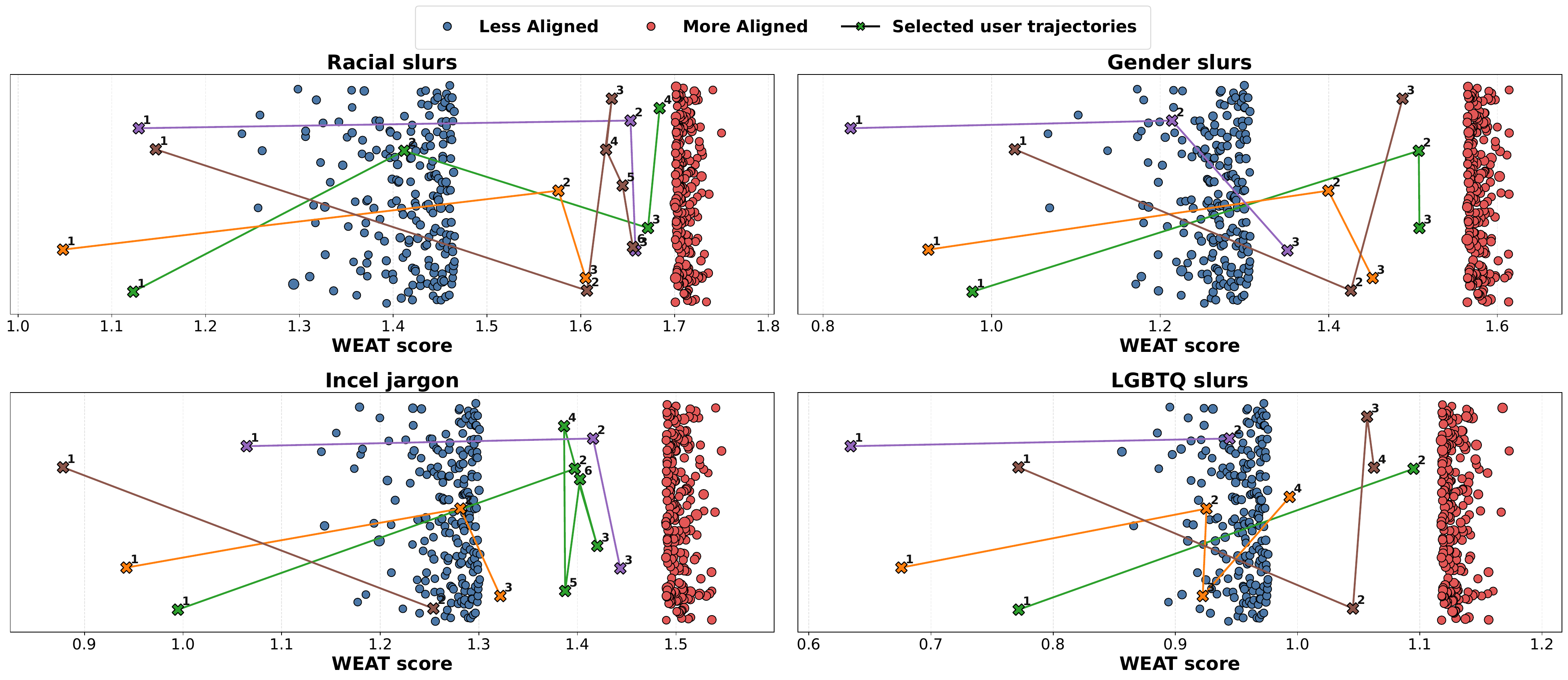}
    \caption{Stormfront dynamics: per-user scatterplots mark least- (blue) and most-aligned (red) accounts on four slur axes; coloured trajectories track selected users with the fastest drift toward alignment.}
    \label{fig:forum-trajectories}
\end{figure*}

Figure~\ref{fig:forum-trajectories} illustrates how selected users’ POLAR scores change as their posting histories grow.
For each highlighted author, we recompute a cumulative association score after the $t$-th post, denoted $s_t(u;\mathcal{A},\mathcal{B})$, using all posts observed up to index $t$.
Each panel plots these cumulative scores on the x-axis (labeled “WEAT score” for historical continuity with prior association-test literature), while the y-axis is an uninformative jitter used only to reduce overlap and make individual steps visible.
Colored markers correspond to successive posting steps (numbers indicate $t$), and connecting lines are drawn solely to guide the eye through the temporal order.

The highlighted trajectories drift steadily rightward across all four axes, indicating that continued participation is associated with increasing alignment between user embeddings and the slur lexicons. This pattern is observational (not causal) but consistent with progressive radicalization signals accumulating in users’ text histories. Because POLAR is lightweight, label-free, and recomputable after each message, these rightward drifts offer an interpretable early-warning indicator -- even for short or sparsely annotated histories.

\section{Conclusion}\label{sec:conclusion}

This study introduced \emph{POLAR}, the first \emph{per-author} lexical association test that operates directly in the embedding space of a lightly adapted masked-language model.  By hashing user identifiers into private tokens and fine-tuning only those token vectors, we generate compact, interpretable representations that remain geometrically compatible with ordinary wordpieces.  Standardized effect sizes computed between each user vector and curated attribute sets then provide a scale-stable, author-level measure of stance or bias -- complete with permutation $p$-values and false-discovery control.

Empirical results across two contrasting corpora underscore the method’s versatility. On the fox8–23 bot–human benchmark, POLAR cleanly separates LLM-driven bots from organic accounts along multiple stylistic axes without ever observing class labels. In the Stormfront forum, the same statistic exposes strong, uneven alignment with hateful or derogatory lexicons, and cumulative trajectories signal progressive radicalization over time. Taken together, these findings show that user embeddings capture stable lexical preferences, that these preferences can be probed with lightweight, modular tests, and that the resulting scores offer immediate, interpretable diagnostics for computational social science.

\section{Limitations and Future Work}\label{sec:limits}

\paragraph{Scope of corpora.}
Our evaluation spans Twitter and a single extremist forum; generalising to other platforms (e.g., long-form blogs, chat applications) or multilingual settings remains future work.

\paragraph{Short histories and silent users.}
POLAR requires at least two posts per author; users with very sparse histories are discarded.  Alternative smoothing or hierarchical pooling strategies could extend coverage.

\paragraph{Attribute design bias.}
Choice of attribute sets steers interpretation.  Although our pairs are drawn from prior literature and domain expertise, they may omit salient frames or inadvertently encode researcher bias.  Crowdsourced or community-generated attributes are a promising mitigation.

\section{Ethical Statement}
We use only public, researcher-released datasets (fox8–23; Stormfront) under their licenses -- no new collection or user interaction. Analyses rely on pseudonymous IDs and irreversible hashed user tokens; we release no account mappings or additional personal data. Sensitive lexicons are used solely for measurement; quotes are minimized with content warnings; results are aggregate only. Code is research-only and explicitly forbids surveillance/profiling uses. This observational public-data study did not require human-subjects review; we follow ICWSM ethics guidelines and discuss risks and limitations.

\section*{Acknowledgments}
This work was partially funded by CNPq, CAPES, FAPEMIG, and IAIA - INCT on AI.

% ------------------------------------------------------------------
% Reference links for works not in your proposal (to fetch BibTeX):
% WEAT (Caliskan et al., Science 2017): https://www.science.org/doi/10.1126/science.aal4230
% SEAT (May et al., NAACL 2019): https://arxiv.org/abs/1903.10561
% Goldfarb-Tarrant et al. (ACL 2021): https://aclanthology.org/2021.acl-long.81/
% UEM (Doddapaneni et al., 2024): https://arxiv.org/abs/2401.04858
% UQABench (KDD 2025): https://dl.acm.org/doi/10.1145/3711896.3737385
% fox8-23 data (Zenodo): https://zenodo.org/records/8035290
% fox8-23 paper: https://arxiv.org/abs/2307.16336
% ------------------------------------------------------------------

% \bibliographystyle{aaai}
\bibliography{aaai2026} % (Add your .bib in the camera-ready version)

\section{Paper Checklist}

\begin{enumerate}

\item For most authors...
\begin{enumerate}
    \item  Would answering this research question advance science without violating social contracts, such as violating privacy norms, perpetuating unfair profiling, exacerbating the socio-economic divide, or implying disrespect to societies or cultures?
    \answerYes{Yes.}
  \item Do your main claims in the abstract and introduction accurately reflect the paper's contributions and scope?
    \answerYes{Yes.}
   \item Do you clarify how the proposed methodological approach is appropriate for the claims made? 
    \answerYes{Yes.}
   \item Do you clarify what are possible artifacts in the data used, given population-specific distributions?
    \answerYes{Yes.}
  \item Did you describe the limitations of your work?
    \answerYes{Yes.}
  \item Did you discuss any potential negative societal impacts of your work?
    \answerYes{Yes.}
      \item Did you discuss any potential misuse of your work?
    \answerYes{Yes.}
    \item Did you describe steps taken to prevent or mitigate potential negative outcomes of the research, such as data and model documentation, data anonymization, responsible release, access control, and the reproducibility of findings?
    \answerYes{Yes.}
  \item Have you read the ethics review guidelines and ensured that your paper conforms to them?
    \answerYes{Yes.}
\end{enumerate}

\item Additionally, if your study involves hypotheses testing...
\begin{enumerate}
  \item Did you clearly state the assumptions underlying all theoretical results?
    \answerYes{Yes.}
  \item Have you provided justifications for all theoretical results?
    \answerYes{Yes.}
  \item Did you discuss competing hypotheses or theories that might challenge or complement your theoretical results?
    \answerYes{Yes.}
  \item Have you considered alternative mechanisms or explanations that might account for the same outcomes observed in your study?
    \answerYes{Yes.}
  \item Did you address potential biases or limitations in your theoretical framework?
    \answerYes{Yes.}
  \item Have you related your theoretical results to the existing literature in social science?
    \answerYes{Yes.}
  \item Did you discuss the implications of your theoretical results for policy, practice, or further research in the social science domain?
    \answerYes{Yes.}
\end{enumerate}

\item Additionally, if you are including theoretical proofs...
\begin{enumerate}
  \item Did you state the full set of assumptions of all theoretical results?
    \answerNA{NA.}
	\item Did you include complete proofs of all theoretical results?
    \answerNA{NA.}
\end{enumerate}

\item Additionally, if you ran machine learning experiments...
\begin{enumerate}
  \item Did you include the code, data, and instructions needed to reproduce the main experimental results (either in the supplemental material or as a URL)?
    \answerYes{Yes.}
  \item Did you specify all the training details (e.g., data splits, hyperparameters, how they were chosen)?
    \answerYes{Yes.}
     \item Did you report error bars (e.g., with respect to the random seed after running experiments multiple times)?
    \answerYes{Yes.}
	\item Did you include the total amount of compute and the type of resources used (e.g., type of GPUs, internal cluster, or cloud provider)?
    \answerYes{Yes.}
     \item Do you justify how the proposed evaluation is sufficient and appropriate to the claims made? 
    \answerYes{Yes.}
     \item Do you discuss what is ``the cost`` of misclassification and fault (in)tolerance?
    \answerYes{Yes.}
  
\end{enumerate}

\item Additionally, if you are using existing assets (e.g., code, data, models) or curating/releasing new assets, \textbf{without compromising anonymity}...
\begin{enumerate}
  \item If your work uses existing assets, did you cite the creators?
    \answerYes{Yes}
  \item Did you mention the license of the assets?
    \answerYes{Yes}
  \item Did you include any new assets in the supplemental material or as a URL?
    \answerNo{No}
  \item Did you discuss whether and how consent was obtained from people whose data you're using/curating?
    \answerYes{Yes}
  \item Did you discuss whether the data you are using/curating contains personally identifiable information or offensive content?
    \answerYes{Yes}
\item If you are curating or releasing new datasets, did you discuss how you intend to make your datasets FAIR (see \citet{fair})?
\answerNA{NA.}
\item If you are curating or releasing new datasets, did you create a Datasheet for the Dataset (see \citet{gebru2021datasheets})? 
\answerNA{NA.}
\end{enumerate}

\item Additionally, if you used crowdsourcing or conducted research with human subjects, \textbf{without compromising anonymity}...
\begin{enumerate}
  \item Did you include the full text of instructions given to participants and screenshots?
    \answerNA{NA.}
  \item Did you describe any potential participant risks, with mentions of Institutional Review Board (IRB) approvals?
    \answerNA{NA.}
  \item Did you include the estimated hourly wage paid to participants and the total amount spent on participant compensation?
    \answerNA{NA.}
   \item Did you discuss how data is stored, shared, and deidentified?
   \answerNA{NA.}
\end{enumerate}

\end{enumerate}

\appendix
\section{Appendix}
\label{sec:appendix}

\section{Artifacts \& File Schema}
\label{sec:app-artifacts}
We release one run folder per corpus:
\begin{itemize}
\item \textbf{fox8–23 (bots vs.\ humans)}: \texttt{out\_bot\_human\_bert\_space/}
\item \textbf{Stormfront (white-supremacist forum)}: \texttt{out\_hate\_bert\_space/}
\end{itemize}
Each contains:
\begin{itemize}
\item \texttt{model/}: HuggingFace tokenizer and embeddings (includes user tokens).
\item \texttt{users.csv}: columns \texttt{user\_id}, \texttt{n\_posts}; optional \texttt{token}, \texttt{label\_majority}, \texttt{targets}.
\item \texttt{meta.json}: run metadata (e.g., \texttt{usr\_prefix}, \texttt{min\_posts\_per\_user}, seed, training flags).
\item \texttt{per\_user\_scores.csv} (long format): \texttt{user\_id}, \texttt{pair}, \texttt{s}, \texttt{p\_perm}, \texttt{n\_posts}, \texttt{n\_pos\_attr}, \texttt{n\_neg\_attr}, \texttt{label\_majority}, \texttt{targets}, \texttt{signif\_bh\_fdr\_0.05}.
\end{itemize}

\section{Attribute Sets}
\label{sec:app-attrs}
We embed exact word/phrase lists directly from the trained vocabulary and publish them with the code/artifacts. Examples (only to aid figure reading):
\paragraph{fox8–23 (register axes).}
\emph{Newswire vs.\ Personal Experience} (e.g., ``court, officials'' \emph{vs.} ``tbh, lol''); 
\emph{Promo/CTA vs.\ Hedges} (``offer, discount'' \emph{vs.} ``I think, not sure'');
\emph{Toxicity vs.\ Civility} (``idiot, garbage'' \emph{vs.} ``thank you, empathy'');
\emph{Scam/Finance vs.\ Daily Life} (``airdrop, seed phrase'' \emph{vs.} ``movie night, weekend'').
\paragraph{Stormfront (sensitive axes).}
\emph{Sentiment}; \emph{Toxicity vs.\ Civility};
\emph{Racial slurs vs.\ Neutral};
\emph{Gender slurs vs.\ Respect};
\emph{LGBTQ slurs vs.\ Neutral};
\emph{Violence vs.\ Peace};
\emph{Incel jargon vs.\ Relationships}.
\smallskip\newline
\textit{Sensitive lexicons are used strictly for measurement on public datasets; outputs are anonymized at analysis time.}

\section{Diagnostics \& Edge Cases}
\label{sec:app-diagnostics}
\begin{itemize}
\item \textbf{Coverage}: per (user, pair) we record how many attribute items survive tokenization; low coverage is flagged in \texttt{per\_user\_scores.csv}.
\item \textbf{Axis sanity}: we log $\cos(\bar{\mathbf a},\bar{\mathbf b})$ (centroid similarity) to detect poorly separated pairs.
\item \textbf{Degeneracy}: if $\operatorname{sd}([d_A;d_B])=0$ or one side is empty post-tokenization, we set \texttt{s}/\texttt{p\_perm}=\texttt{NaN}; these remain non-significant after FDR.
\item \textbf{Missing user tokens}: users whose tokens are absent from the saved vocabulary are skipped and counted at inference time.
\end{itemize}

\section{Compute \& Scaling}
\label{sec:app-compute}
Let $U$ be users, $d$ the embedding dimension, and $K$ total attributes per pair. Inference costs $O(UKd)$ for dot products; each permutation replicate operates in $O(K)$ (relabel only). With $M{=}2000$ and modest $K$ (tens), runtime grows linearly with $U$ and the number of tested pairs.

\section{Training \& Hyperparameters}
\label{sec:app-train}
This section centralizes training settings to avoid cluttering Methodology.
\paragraph{Base LM \& user tokens.}
\texttt{bert-base-uncased} (HuggingFace). One deterministic private token per user, format \texttt{usr<sha1[:10]>}, \emph{prepended} to every post. Vocabulary resized once to include all user tokens. Users with $<$2 posts are excluded.
\paragraph{Objective.}
Masked-LM cross-entropy with mask prob.\ 0.15. Additional user-token mask prob.\ 0.30. Alignment term that nudges the user token toward the mean of non-user token embeddings in the post; weight $0.2$.
\paragraph{Batching.}
Batches are approximately balanced across users. Optional per-epoch cap of posts per user (default: off). Values are recorded in \texttt{meta.json}.
\paragraph{Optimizer \& schedule.}
AdamW (betas $[0.9,0.999]$, $\epsilon=10^{-8}$, weight decay $0.01$). Learning rate $5{\times}10^{-5}$ with linear warm-up (ratio $0.03$) then linear decay. Gradient clip $1.0$.
\paragraph{Run settings.}
Sequence length $128$; batch size $128$; epochs $4$; gradient accumulation $1$; dropout as in BERT-base defaults. Mixed precision: bf16 when available, else fp16. Random seed $123$ (NumPy/PyTorch). No early stopping; final checkpoint is used.
\paragraph{Stability.}
Short freeze–then–unfreeze schedule for user-token rows during early steps (stabilizes token learning without changing downstream inference).
\paragraph{Compute.}
Single-GPU training; inference is CPU/GPU agnostic.
\paragraph{Provenance.}
Runs write \texttt{meta.json} (hyperparameters, \texttt{usr\_prefix}, \texttt{min\_posts\_per\_user}) and save the tokenizer with user tokens under \texttt{model/}.

\end{document}